%% file: main.tex
\documentclass[10pt,twocolumn,letterpaper]{article}
\usepackage{amsfonts}

\usepackage[pagenumbers]{cvpr} 

\input{preamble}

\definecolor{cvprblue}{rgb}{0.21,0.49,0.74}
\usepackage[pagebackref,breaklinks,colorlinks,citecolor=cvprblue]{hyperref}
\usepackage{amsmath}
\usepackage{multirow}
\usepackage{makecell}
\usepackage{pifont}
\newcommand{\cmark}{\ding{51}}%
\newcommand{\xmark}{\ding{55}}%
\usepackage{dblfloatfix}
\usepackage[bottom]{footmisc}

\def\paperID{7672} 
\def\confName{CVPR}
\def\confYear{2024}

\title{VoxelKP: A Voxel-based Network Architecture for Human Keypoint Estimation in LiDAR Data}

\author{Jian Shi, Peter Wonka\\
King Abdullah University of Science and Technology\\
{\tt\small \{jian.shi, peter.wonka\}@kaust.edu.sa}
}

\begin{document}
\maketitle
\begin{abstract}

We present \textit{VoxelKP}, a novel fully sparse network architecture tailored for human keypoint estimation in LiDAR data.
The key challenge is that objects are distributed sparsely in 3D space, while human keypoint detection requires detailed local information wherever humans are present.
We propose four novel ideas in this paper.
First, we propose sparse selective kernels to capture multi-scale context. Second, we introduce sparse box-attention to focus on learning spatial correlations between keypoints within each human instance. Third, we incorporate a spatial encoding to leverage absolute 3D coordinates when projecting 3D voxels to a 2D grid encoding a bird's eye view.
Finally, we propose hybrid feature learning to combine the processing of per-voxel features with sparse convolution. We evaluate our method on the Waymo dataset and achieve an improvement of $27\%$ on the MPJPE metric compared to the state-of-the-art, \textit{HUM3DIL}, trained on the same data, and $12\%$ against the state-of-the-art, \textit{GC-KPL}, pretrained on a $25\times$ larger dataset.
To the best of our knowledge, \textit{VoxelKP} is the first single-staged, fully sparse network that is specifically designed for addressing the challenging task of 3D keypoint estimation from LiDAR data, achieving state-of-the-art performances. Our code is available at \url{https://github.com/shijianjian/VoxelKP}.
\end{abstract}


\section{Introduction}
\label{sec:intro}

Human pose estimation is a critical area of research with applications spanning computer vision, robotics, human-computer interaction, and augmented/virtual reality. 
Previous works~\cite{toshev2014deeppose,newell2016stacked,sun2019deep} are mostly based on 2D images and videos. 
Compared to regular RGB input, LiDAR sensors provide detailed 3D structural information by measuring the distance to objects using laser light.
Apart from its robustness under occlusion and illumination changes, LiDAR also offers privacy protection as it can not retain facial details.
In recent years, significant progress has been made in 3D object detection from LiDAR point clouds, with methods like PointRCNN~\cite{shi2019pointrcnn}, Part-A2~\cite{shi2019part}, and PV-RCNN~\cite{shi2020pv} achieving impressive results, while human pose estimation from LiDAR is still an open research problem with much room for improvement.
Typically, object detection methods focus on capturing objects scattered sparsely across the 3D space while the keypoints tend to be distributed densely within localized regions around the human body. 
This fundamental discrepancy in the context captured by existing detectors limits their suitability for precise 3D keypoint prediction due to the lack of fine-grained spatial information.
To address this gap, we aim to extend the success of 3D object detection to 3D keypoint estimation for Lidar point cloud data by introducing novel components to preserve fine-grained spatial information.

This work identifies the importance of learning local dense features to capture the intricate spatial relationships between keypoints for precise human pose estimation. For this purpose, we introduce the \textit{VoxelKP} architecture. \textit{VoxelKP} is a novel, fully sparse neural network tailored specifically for human keypoint estimation within LiDAR point clouds. It combines local feature extraction and global context modeling to achieve accurate human pose prediction from LiDAR scans.
To be specific, we introduce four key components that play a pivotal role in enhancing local feature learning for keypoint estimation:
\begin{itemize}
    \item \textbf{Sparse Selective Kernel (SSK) Modules}: These modules are designed to selectively aggregate multi-scale 3D features efficiently extracted at sparse voxel locations. By employing various receptive field kernels and a selection mechanism, the SSK modules significantly improve spatial context. This is crucial for the accurate estimation of keypoints, as it allows the model to understand the spatial relationships between keypoint locations.
    \item \textbf{Sparse Box-Attention Modules}: Our approach incorporates localized box-based self-attention to partition the sparse voxel space into non-overlapping box regions. This strategy enables the model to capture dependencies between voxels within each box. By doing so, it extracts fine-grained local features necessary for resolving densely distributed keypoints. This focused modeling of intricate spatial relationships between keypoints is instrumental in achieving precise human pose estimation.
    \item \textbf{Spatially Aware Multi-Scale BEV Fusion}: To retain the 3D spatial relationships between keypoints, we introduce a spatially aware multi-scale bird's eye view (BEV) fusion technique. This innovative approach encapsulates 3D spatial information into 2D representations, thereby improving the accuracy of keypoint estimation. It ensures that the model considers spatial information when predicting keypoints, enhancing the overall performance.
    \item \textbf{Hybrid Feature Learning} In addition to the above components, we propose the use of hybrid feature learning. We combine the results of two parallel branches in the architecture: per-voxel computations using MLPs and sparse convolutions that process voxel-neighborhoods.
\end{itemize}

To the best of our knowledge, \textit{VoxelKP} is the first single-staged, fully sparse network that is specifically designed for addressing the challenging task of 3D keypoint estimation from LiDAR data, achieving $27\%$ on the MPJPE metric compared to the current state-of-the-art trained on the same data.

\section{Related Work}

\subsection{Deep Learning on Point Clouds}

Many neural network architectures have been adapted for processing point clouds. Earlier methods like VoxNet~\cite{maturana2015voxnet} applied 3D CNNs to voxel grids for object classification.
PointNet~\cite{qi2017pointnet} was one of the first works to operate directly on point clouds using MLPs and max pooling to extract global features of entire scenes represented by point clouds. Follow-up works like PointNet++~\cite{qi2017pointnet++} introduced hierarchical and localized feature learning. Meanwhile, another branch of works such as PointCNN~\cite{li2018pointcnn} and KPConv~\cite{thomas2019kpconv} introduced novel convolutional operators for learning features on the unordered point clouds, overcoming the limitations of typical convolutions for this irregular data type.

Typical LiDAR-generated point clouds contain more than $100,000$ points, making point-by-point computations overwhelming due to the massive data scale.
VoxelNet~\cite{Zhou_2018_CVPR} proposed a voxel feature encoding (VFE) layer as a workaround for the high computational and memory issues brought by point-by-point computations.
Meanwhile, sparse and submanifold sparse convolution operations~\cite{graham20183d} exploit sparsity in the voxel grid to reduce computations. SECOND~\cite{yan2018second} introduced an efficient sparse convolutional approach that benefits from the sparse operations.
Following SECOND, subsequent works like PointPillars \cite{lang2019pointpillars}, 3DSSD~\cite{yang20203dssd}, PV-RCNN~\cite{shi2020pv}, CenterPoint~\cite{yin2021center} further advanced sparse convolutional detection on point clouds, introducing ideas like pillar encoding for faster detection, multi-scale detection stacks with anchor boxes, shared voxel encoders, and detecting small objects by center points. VoxelNeXt~\cite{chen2023voxenext} further demonstrates a fully sparse voxel-based method without sparse-to-dense conversion or NMS post-processing.
However, these approaches are targeted at improving bounding box localization accuracy, which does not require fine-grained spatial features for precise keypoint estimation tasks. Instead, We propose \textit{VoxelKP}, a novel sparse convolutional architecture tailored for learning discriminative local features from sparse LiDAR data for accurate human pose estimation.

\subsection{Human Pose Estimation on Point Clouds}

Human pose estimation has been extensively studied in images, with methods like DeepPose~\cite{toshev2014deeppose}, Stacked Hourglass~\cite{newell2016stacked}, and HRNet~\cite{sun2019deep} achieving high accuracy on benchmarks like COCO-wholebody~\cite{jin2020whole}. However, compared to RGB images, point clouds provide explicit 3D structural information about the shape and depth of objects.
Shotton \etal~\cite{shotton2011real} pioneered point cloud human pose estimation from a single depth image. Recent works such as~\cite{zhou2020learning,ma2021context} proposed a deep learning-based 3D human pose estimation from depth images. 
Waymo~\cite{Sun_2020_CVPR} has released keypoint annotations for LiDAR-collected point cloud scenes, while only $3\%$ of the frames are annotated with keypoint human poses. 
Due to the scarcity of the keypoint annotations within LiDAR point cloud data, many works have taken semi-supervised or weak-supervised approaches to compensate for the limited availability of labeled 3D pose data.
For example, some works~\cite{zanfir2023hum3dil,zheng2022multi} took a multi-modal approach to utilize the enriched image annotations to assist the recognition from point clouds. Weng \etal~\cite{weng20233d} proposed an unsupervised approach that generates pseudo ground truth without using annotated keypoint data, along with a fine-tuning approach that pretrains the model with synthetic data then fine-tunes on the training set.
A concurrent work~\cite{ye2023lpformer} adopted a fine-tuning strategy that used a frozen backbone pretrained on a large-scale dataset as a feature extractor, achieving plausible performance.
In general, multi-person pose estimation from sorely point clouds remains relatively unexplored due to the lack of ground-truth 3D human pose annotations.
This work proposes a single-staged keypoint estimation method with only LiDAR point clouds, achieving comparable performances without extra training data.

\section{Method}

LiDAR point clouds typically contain sparsely distributed objects that occupy only small regions of the full 3D space. While the distribution of humans in space is sparse, in contrast, human keypoints require dense information wherever a human is present. To handle this density variation, we aim to improve feature learning in the regions where keypoints need to be located and detailed information is required.
In this section, we first present the formulation of the task, then introduce the key components proposed in our network, and finally elaborate on the details of the network architecture.

\subsection{Problem Formulation}

Given a 3D point cloud scanned by LiDAR sensors, our goal is to estimate the 3D locations of $K$ keypoints that represent the human pose. Let the input point cloud $P$ be $\mathbb{R}^{N\times C}$ where $N$ is the number of points and $C$ is the number of features (\eg x, y, z, intensity, elongation).
We use a sparse voxel representation to represent point clouds, which consists of two separate tensors: one feature tensor $\mathbb{R}^{V\times C}$ and one index tensor $\mathbb{R}^{V\times 4}$ where $V$ is the number of non-empty voxels and $4$ dimensions are used for batch sample index and the three coordinates of each voxel.
We define the ground truth pose for the $i^{th}$ human as a set of 3D keypoint locations $G_i = \{g_i^1, g_i^2, ..., g_i^K\}$ where $g_i^k \in \mathbb{R}^3$ is the location of the $k^{th}$ keypoint in the global coordinate frame. The set of $K$ keypoints corresponds to anatomical joints of interest such as shoulders, elbows, wrists, hips, knees, and ankles. Our objective is to predict the 3D keypoint locations from the input point cloud, i.e. to learn a function F such that $\hat{G} = F(P)$, where $\hat{G} \in \mathbb{R}^{M\times K\times3}$ is the tensor of predicted 3D keypoint locations of $M$ humans.


\subsection{Key Components}

See ~\cref{fig:main} for our final architecture \textit{VoxelKP}. First, the scene is voxelized into a sparse 3D grid. Then the sparse grid goes through multiple 3D blocks to extract multi-scale 3D sparse features, followed by a projection into a sparse 2D grid, and 2D blocks. Finally, multiple prediction heads output the keypoints.  
The proposed architecture contains four key components for enhancing the spatial localization accuracy of keypoints.
Specifically, we employ spatially aware \textit{sparse selective kernel modules} and \textit{sparse box-attention modules} in our network to improve the representational power to encode and localize the fine-grained keypoint features.
In addition, we use a \textit{spatially aware multi-scale BEV fusion} method to encode the spatial information, along with a multi-scale fusion to understand the context across varying densities.
Lastly, we use a \textit{hybrid feature learning} approach to capture both fine-grained per-voxel details and relatively coarse-grained local neighborhood information.

\begin{figure}[h]
    \centering
    \includegraphics[width=\linewidth]{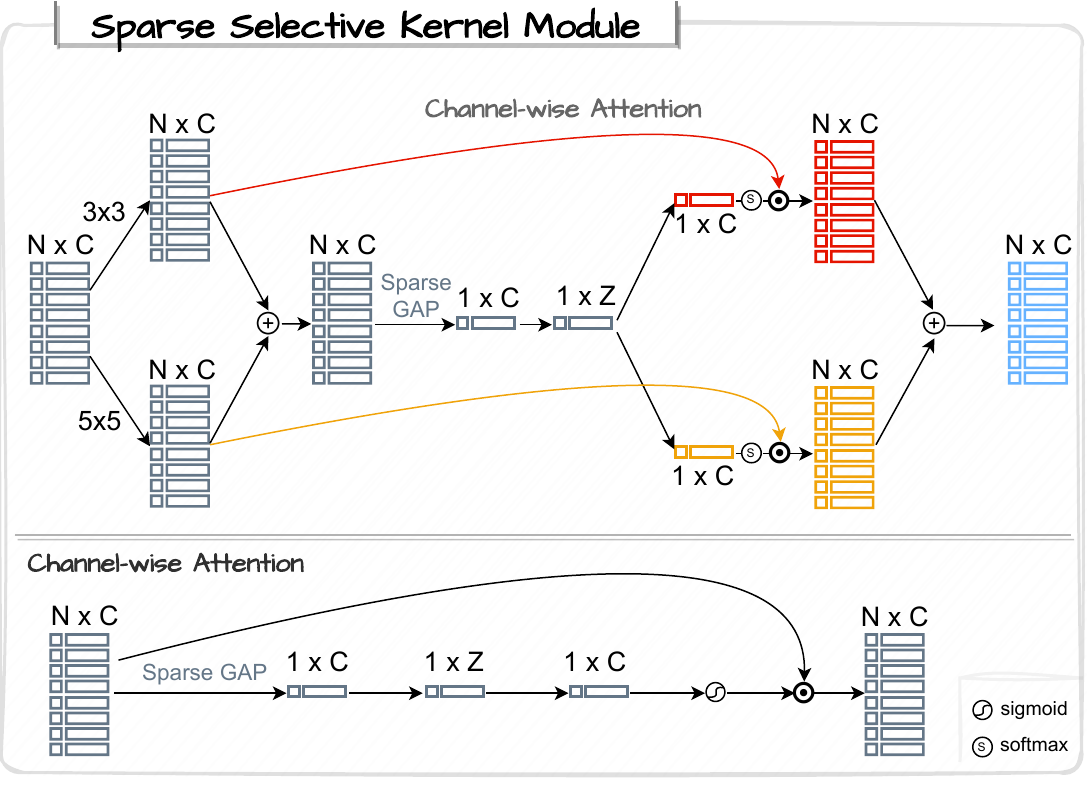}
    \caption{Sparse selective kernel module with one sample input. The SSK module selects the best kernels from different receptive fields with a softmax-based channel-wise attention mechanism.}
    \label{fig:ssk}
\end{figure}

\subsubsection{Sparse Selective Kernel Module}

Inspired by~\cite{hu2018squeeze,li2019selective}, we propose the sparse selective kernel (SSK) module that selectively aggregates multi-scale features to improve spatial context.
The SSK modules perform spatial attention on a 3D sparse voxel space, where the attention specializes the receptive field at each position using a data-driven kernel selection.
As demonstrated in~\cref{fig:ssk}, we first generate a set of sparse 3D submanifold convolution kernels with varied receptive field sizes of $3\times 3\times 3$ and $5\times 5\times 5$~
. A submanifold convolution computes output values only if the convolution kernel is centered on a non-empty voxel, i.e., the number of non-empty voxels remains the same. These operations are applied to sparsely sampled voxel locations, extracting multi-scale features while remaining efficient.
Next, the features from each kernel are fed into a selection module that compresses the spatial dimension by a global average pooling (GAP), and then a feature squeeze and expansion are applied. In our implementation, $Z$ is 25\% of $C$. The resulting tensor would then be used to weigh the features after a softmax activation. 
We denote a voxel position as $p$, the set of voxel positions  within a voxel grid as $P_s$, and the feature corresponding to voxel position $p$ as $f_p$. The sparse GAP $\Bar{F}_s$ can be obtained by:
\begin{gather}
    \Bar{P}_s=\{(x_p,y_p,z_p) | p \in P_s \}, \nonumber \\
    \Bar{F}_s=\{\frac{1}{|S_{\Bar{P}}|}\sum_{p\in S_{\Bar{P}}} f_p | \Bar{p} \in \Bar{P}_s \},
\end{gather}
where $|S_{\Bar{P}}|$ is the number of valid voxels for the sample $s$ in a batch.
This produces channel-wise attention weights, allowing the network to selectively emphasize or suppress each kernel's features. The multi-scale local features can then be obtained by combining weighted features from all kernels through averaging.

\begin{figure}[t]
    \centering
    \includegraphics[width=\linewidth]{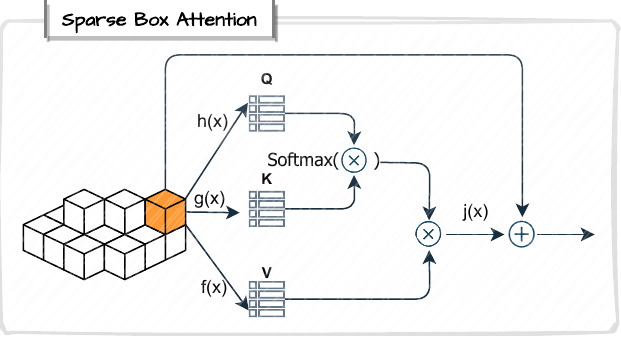}
    \caption{Sparse box-attention module. This attention mechanism selects the voxel features that correspond to one box partition referring to the index tensor and then performs self-attention on the selected voxels. The functions f,g,h, and j are linear layers.}
    \label{fig:spa}
\end{figure}

\subsubsection{Sparse Box-Attention Module}
We apply box-based self-attention. Unlike the previous works that tried to capture a wider range of global features with self-attention methods for segmentation tasks~\cite{lai2022stratified,lai2023spherical}, we focus on local feature extraction to resolve the densely distributed keypoints in local regions. The key idea is to partition the sparse 3D voxel space into non-overlapping boxes. Within each local box, we apply self-attention to capture dependencies between the voxels inside the box.
The features in each box go through a linear layer for the queries $Q$, keys $K$, and values $V$, where $Q,K,V\in \mathbb{R}^{k_{b}\times h \times d}$ and $k_b,h,d$ are the number of valid voxels in the $b$-th box, attention heads, and feature dimensions. Since we are using sparse tensor representations, each box partition may contain a varying number of voxels. Referring to~\cite{lai2022stratified}, we then compute the attention map by the following equation:
\begin{gather}
    \text{Attention}_{i,h}=\sum_{j=1}^{k_b} \text{softmax}(Q_{i,h}\cdot K_{j,h}) \times V_{j,h}.
\end{gather}
We then further apply an additional projection layer on the obtained attention map, as shown in~\cref{fig:spa}.

\subsubsection{Spatially Aware Multi-Scale BEV Fusion}

\begin{figure}[b]
    \centering
    \includegraphics[width=\linewidth]{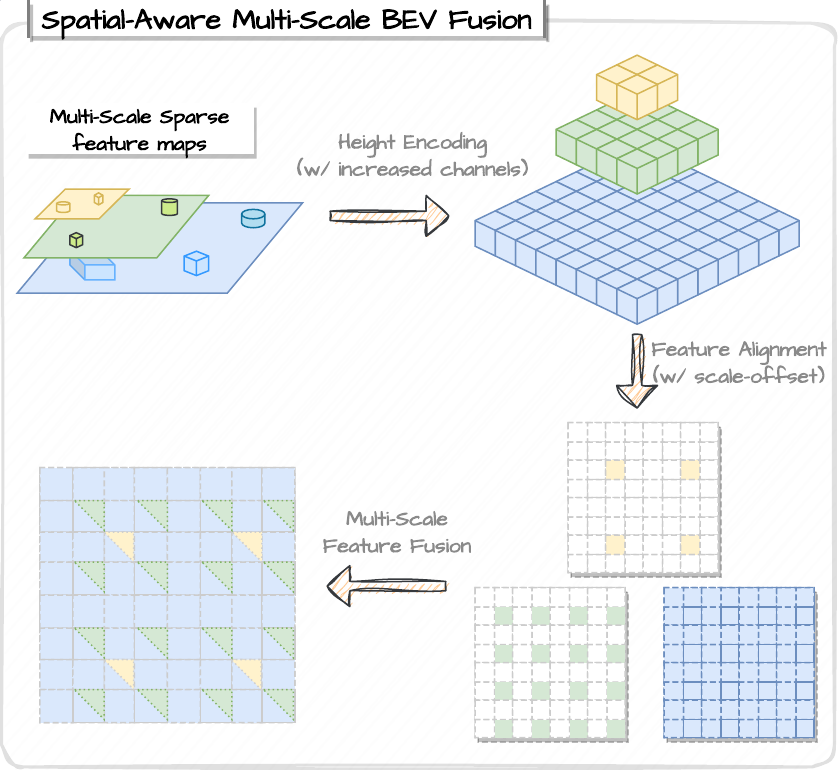}
    \caption{Spatially aware multi-scale BEV Fusion module. Note that we use a dense representation for a better visual illustration of the method. }
    \label{fig:bev}
\end{figure}

Compressing features into bird's eye view (BEV) maps is a common practice for object detection~\cite{chen2017multi,yan2018second}.
For a sparse 3D voxel grid of size $C \times X \times Y \times Z$, we use $C$ to denote the number of features per voxel, $X$ and $Y$ as the spatial extent in the ground plane, and $Z$ as the up axis.
Starting with a sparse 3D voxel grid, previous works such as~\cite{chen2023voxenext} simply ignore the height information by summing the features of all voxels that share the same position on the ground plane (the same $x$ and $y$ coordinates). However, different from object detection tasks, height information is essential for keypoint estimation tasks to precisely locate each keypoint.
A reasonable approach is to directly deploy 3D feature maps. Unfortunately, this direct 3D approach does not lead to a decent performance as training does not converge well, as shown in~\cref{tab:abl_bev}.
We, therefore, propose a simple \textit{spatially aware multi-scale BEV fusion} approach for fusing features from multiple encoder layers in a way that retains spatial information, as illustrated in~\cref{fig:bev}.


\begin{figure*}[t]
    \centering
    \includegraphics[width=\linewidth]{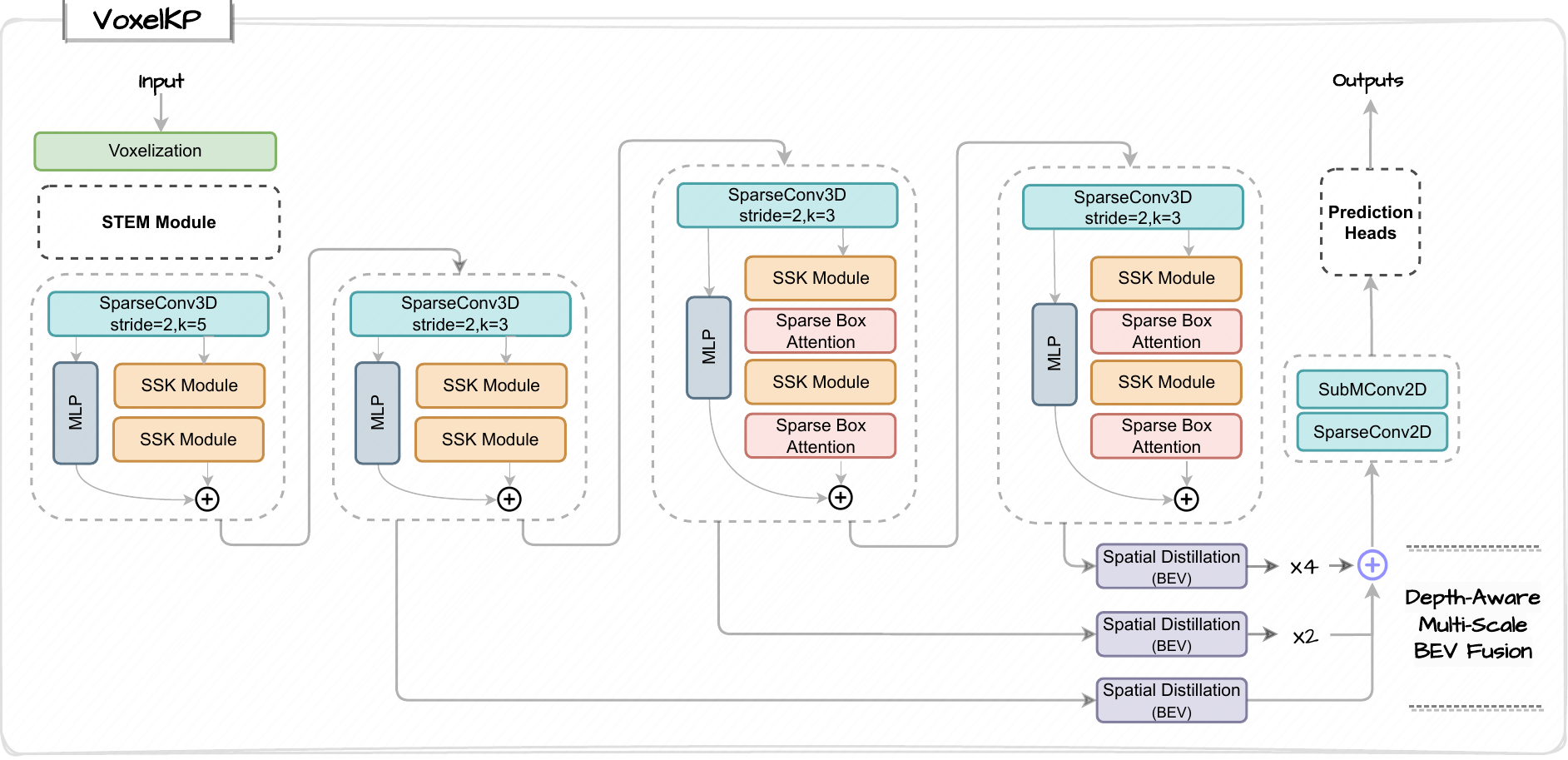}
    \caption{The overall architecture of the \textit{VoxelKP}.}
    \label{fig:main}
\end{figure*}

\noindent\textbf{Height Encoding} Transforming 3D data into BEV is often used in 3D object detection and segmentation tasks, for reducing the dimensionality of point clouds and making them more manageable for processing. An object detection method may project the 3D voxel grid to a 2D BEV representation by adding features from voxels that share the same x and y position, losing the information about which height a feature was taken from. Instead, we use a height encoding method. Specifically, we compress the height dimension to 1 using convolution kernels of size $(1, 1, h)$ where $h$ is the height of each 3D voxel grid. Meanwhile, we increase the number of resulting channels to retain more spatial details and features from the 3D representation. 
This provides a richer representation for the 2D regression heads to work with.

\noindent\textbf{Multi-scale Feature Fusion}
After obtaining $z$ multi-scale height-encoded BEV maps from the last few stages of the network, we then fuse those feature maps to create a feature map that contains multi-scale features. Unlike working with dense tensors, the direct interpolation of the feature maps in the sparse case is computationally complex, as it requires specialized algorithms to efficiently navigate through the predominantly empty voxels to find and interpolate the adjacent non-empty voxels. Instead, we directly modify the feature position of the sparse tensor by multiplying the voxel position by its scale $r$. To avoid overlapping feature position of $(p_x * 2^r, p_y * 2^r)\; r\in\{0,1,2...\}$ during the scale multiplication, we align the xy-plane positions $(p_x,p_y)$ using scale offsets $(p_x * 2^r + r, p_y * 2^r + r)$.

By stacking the $r$-scaled feature maps together, we obtain a multi-scale 3D feature map with a height of $r$.
To obtain a BEV feature map, instead of collapsing with $1\times 1 \times r$ convolutions, we simply apply an intuitive scaling for each scale of the feature map. The scaling factor is proportional to the height (scale) of the 3D feature map. The compressed sparse features $\Bar{F}_c$ and their positions $\Bar{P}_c$ are obtained as:
\begin{gather}
     \Bar{F}=\{f_p \cdot \hat{r}_p | p \in P_c \}, \quad \Bar{P}_c=\{(x_p,y_p) | p \in P_c \}, \nonumber \\
    \Bar{F}_c=\{\sum_{p\in S_{\Bar{p}}} \Bar{f}_p\ | \Bar{p} \in \Bar{P}_c \},
\end{gather}
where $\Bar{F}$ contains the scaled features by the scale offsets and $S_{\Bar{p}}=\{ p|x_p=x_{\Bar{p}}, y_p=y_{\Bar{p}}, p \in P_c\}$ contains voxels that are put onto the same 2D position $\Bar{p}$. $\hat{r}_p$ is the normalized height position of each individual feature.

\subsubsection{Hybrid Feature Learning}

The convolutional operations focus on understanding spatial hierarchies and local geometric structures to extract local neighborhood information.
Concurrently, inspired by the previous point-voxel networks~\cite{liu2019pvcnn,shi2020pv}, we include an MLP branch for each stage. The integration of an MLP branch alongside a convolutional branch is a strategic approach to capture both fine-grained per-voxel details and relatively coarse-grained local neighborhood information.
Each MLP branch is composed of three sequential blocks, each consisting of a linear layer, batch normalization, and a ReLU activation function. The number of channels in each linear layer is set to match the channels of the incoming tensor. We then merge the output features from the MLP and convolutional branches through element-wise summation to create hybrid features of the per-voxel and per-neighborhood information.
This hybrid feature learning approach is deployed to retain and process fine details across the voxel space, which is critical for the accurate localization of keypoints.

\begin{table*}[h]
    \footnotesize
    \newcommand{\greyrule}{\arrayrulecolor{black!30}\cmidrule(lr){1-4}\arrayrulecolor{black}}
    \setlength\tabcolsep{3pt}
    \centering
    \begin{tabular}{p{4cm}|l p{8cm} | c}
        \toprule
        \hspace{1.5cm} Method & Dataset & Description  & MPJPE cm.\\
        \midrule
        \midrule
        \multicolumn{4}{l}{With Extra Training Data} \\
        \greyrule
        \hspace{.3cm} Zheng \etal~\cite{zheng2022multi} \hfill {\color{darkgray} \tiny (CVPR 22)} & Internal dataset + Waymo v.? & Trained on $155,182$ objects from internal data. Generated pseudo labels from 2D image labels. & 10.80 \\
        \hspace{.3cm} GC-KPL \cite{weng20233d} \hfill {\color{darkgray} \tiny (CVPR 23)}  & Waymo v.? &  Pre-trained on synthetic data. Fine-tuned on ground truth & 11.27 \\
          &  Waymo v.? & Pre-trained on $200,000$ Waymo objects. Fine-tuned on ground truth & 10.10 \\
        \midrule
        \midrule
        \multicolumn{4}{l}{Without Extra Training Data} \\
        \greyrule
        \hspace{.3cm} HUM3DIL \cite{zanfir2023hum3dil} \hfill {\color{darkgray} \tiny (CoRL 22)} & Waymo v.1.3.2 & Randomly initialized & 12.21 \\
        \hspace{.3cm} VoxelKP & Waymo v.1.4.2 & Randomly initialized & ~~\textbf{8.87} \\
        \bottomrule
    \end{tabular}
    \caption{Benchmark results. The numbers in the table are taken from their corresponding papers aside from \textit{HUM3DIL}, which is taken from \textit{GC-KP}L paper. It is unclear about the exact training dataset used for \textit{Zheng \etal} and \textit{GC-KPL}. Waymo v1.3.2 and Waymo v1.4.2 share the same data for keypoint estimation task.}
    \label{tab:main}
\end{table*}

\subsection{Network Architecture}

We propose a single-stage, fully sparse neural network, designed for human pose estimation within LiDAR point clouds. The architecture is demonstrated in~\Cref{fig:main}. 
The input is a point cloud $\mathbb{R}^{N\times C}$ where $N$ is the number of points and $C$ is the number of features (\eg x, y, z, intensity). We voxelize the point cloud into a sparse voxel representation.
Our method consists of an input stem network and four stages with gradually decreased feature map size, where each stage reduces the spatial shape of the sparse voxel space by a factor of two. The input stem network is a simple stack of convolution layers, as shown in~\cref{sec:suppl_net}, to extract low-level features from the voxelized point cloud. Next, we apply our proposed SSK modules in our next two stages to better capture the multi-scale local features. We further include window-based self-attention modules for our last two blocks to emphasize local-region features. Note that we do not increase the number of channels for the last three stages.
For each stage, we further include a side MLP branch for learning hybrid features.
We then convert the resulting 3D feature maps from the last three blocks to 2D spatial-encoded BEV representations. Note that we increase the number of channels for the BEV representation to compensate for the information loss of the BEV conversion. These 2D features are further refined with 2D convolutions to aggregate spatial context.
In the end, we obtain the estimated keypoints $Y_{kp}\in \mathbb{R}^{K\times 3}$ and the corresponding predicted visibilities $Y_{kp}\in \mathbb{R}^{K}$, where $K$ is the number of keypoints.


\section{Experiments}

\subsection{Implementation Details}

\textbf{Dataset} We use the Waymo v1.4.2 dataset~\cite{Sun_2020_CVPR}. During the training, we merged ``Pedestrian'' and ``Cyclist'' classes together as a ``Human'' class. Note that there are only $8,125$ human examples with keypoint annotations whilst over 1 million bounding box annotations. We therefore removed the points inside those bounding boxes without keypoint annotations. Each human object is labeled with 14 3D keypoints (nose, left/right shoulders, left/right elbows, left/right wrists, left/right hips, left/right knees, and left/right ankles, head).

\noindent\textbf{Network} The architecture of the network is composed of a stem module followed by four stages, with output channels set to 64, 128, 256, 256, and 256, respectively. Given the high resolution (\eg $1504\times 1504\times 61$) of the voxelized point cloud input, we employ larger sparse convolution kernels (kernel size $k=5$) for the downsampling block in both the stem module and the initial stage. For the subsequent three stages, we revert to a smaller kernel size ($k=3$). To compensate for the information loss in the BEV projection, we increased the channels from 256 to 384 during this process.

\noindent\textbf{Training} We use the point cloud range of the Waymo dataset as $(150.4m, 150.4m, 6m)$ and we transform them into voxel representations by a voxel size of $(0.1m, 0.1m, 0.1m)$. We directly use the global keypoint locations without any encoding. Due to the limited number of training samples, we first apply a ground truth sampling technique~\cite{yan2018second,chen2022autoalignv2}  to concatenate target objects from other frames into the sampled frames. Next, 
we apply global augmentations on the whole point cloud, including random flips on the $x$ and $y$ axes, random scale of the range of $[0.95,1.05]$, and random rotation ranged from $[-\pi/4,\pi/4]$. Additionally, we apply local augmentations on each annotated object, including the random scale of the range of $[0.95,1.05]$, random rotation ranged from $[-\pi/20,\pi/20]$, random frustum dropout~\cite{hu2021pattern} with an intensity range from $[0., 0.2]$, and random noise around the object. Our model is trained using AdamW~\cite{loshchilov2017decoupled} optimizer plus OneCycle~\cite{smith2019super} learning rate scheduler to mitigate overfitting~\cite{smith2018disciplined}.
Specifically, we use a learning rate of $0.003$, weight decay of $0.01$, and $0.9$ momentum. Aside from the regular regression loss and heatmap loss, we include a skeleton regularization loss to make the model aware of the spatial relationships of keypoints.
The details of the used loss functions can be found in~\cref{sec:loss}.

\begin{figure*}
    \centering
    \includegraphics[width=0.95\textwidth]{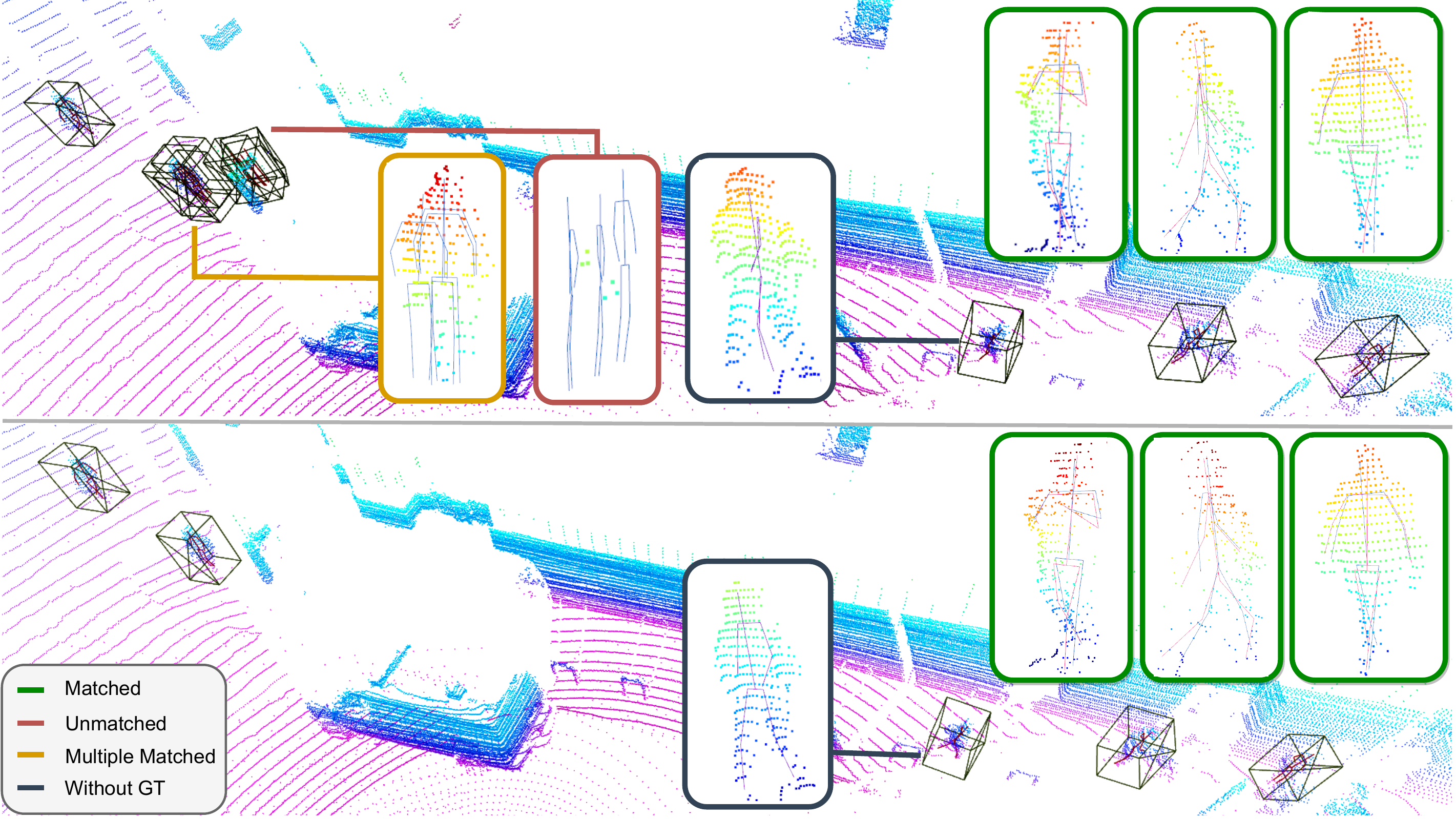}
    \caption{A visual demonstration of our baseline model (top) and the proposed \textit{VoxelKP} (bottom).  Our \textit{VoxelKP} offers improved keypoint estimation with precise locations and fewer false positives. The insets are color-coded according to the legend in the figure. In the green-colored insets, a comparison with the ground truth is shown, with ground truth in red and predictions in blue.}
    \label{tab:visual}
\end{figure*}

\begin{table}[h]
    \footnotesize
    \centering
    \begin{tabular}{l| ccc}
        \toprule
        part  & MPJPE & OKS@AP & PEM \\
        \midrule
        Head      & 0.0570 & 0.6393 &  0.1569 \\
        Shoulders & 0.0669 & 0.8917 &  0.1563 \\
        Elbows    & 0.0948 & 0.7197 &  0.1746 \\
        Wrists    & 0.1467 & 0.3791 &  0.1987  \\
         Hips     & 0.0670 & 0.9533 &  0.1576 \\
        Knees     & 0.0820 & 0.8586 &  0.1660 \\
        Ankles    & 0.1084 & 0.7581 &  0.1765  \\
        \midrule
        All & 0.0887 & 0.7300 & 0.1695 \\
        \bottomrule
    \end{tabular}
    \setlength{\belowcaptionskip}{-1.em}
    \caption{Full evaluation of \textit{VoxelKP}.}
    \label{tab:main_full}
\end{table}

\subsection{Benchmark Methods}

There is a limited number of relevant research for this task. Most of the prior works utilize additional training data beyond the 3D keypoint data within the Waymo dataset. To provide a fair comparison, we need to consider approaches that use extra data and those that rely solely on Waymo ground truth separately. Zheng~\etal ~\cite{zheng2022multi} adopted a pseudo-label generation approach to provide stronger supervision. It utilizes an internal dataset as training data and uses the Waymo dataset for evaluation. \textit{GC-KPL}~\cite{weng20233d} pre-trains its backbone model with extra synthetic or real-world data, then fine-tunes the model with the full Waymo training set.
Given the reliance on extra data in these methods, we consider the LiDAR-only version of \textit{HUM3DIL}~\cite{zanfir2023hum3dil} as our primary competitor. \textit{HUM3DIL} shares the exact same training data as our approach, allowing a direct comparison of techniques.

\subsection{Results}

Previous methods like \textit{GC-KPL} use a subset of the validation data for evaluation, while we evaluate our method with the full validation set for better reproducibility.
We report MPJPE on matched keypoints for our benchmark, following prior works.
As shown in~\cref{tab:main}, we outperform the baseline \textit{HUM3DIL} by approximately $27\%$ in MPJPE.
Our approach achieves state-of-the-art results among methods trained solely on Waymo ground truth. 
We also surpass the approaches leveraging extra synthetic data, beating \textit{Zheng \etal} with synthetic pseudo labels by around $18\%$ and \textit{GC-KPL} with synthetic point clouds by about $21\%$. We achieve better performances as the SOTA \textit{GC-KPL} approach which is pre-trained on $200,000$ real-world samples by about $12\%$. Overall, we demonstrate significant improvements over both the baseline solely using Waymo 3D keypoint data, as well as other techniques relying on extra data. A visual demonstration is presented in~\cref{tab:visual}.

In addition, we report the full spectrum of the evaluation in~\cref{tab:main_full}, including MPJPE, OKS@AP, and PEM. The details for each metric can be found in~\cref{sec:metrics}.

\begin{table*}[]
    \footnotesize
    \centering
    \begin{tabular}{cccc|ccccccccc}
        \toprule
        \multicolumn{4}{c}{Components} & \multicolumn{8}{c}{MPJPE} & PEM \\
        \cmidrule(lr){1-4}\cmidrule(lr){5-12}\cmidrule(lr){13-13}
        Spatial BEV & SSK & Attention & Hybrid Feat.  & head & shoulders & elbows & wrists & hips & knees & ankles & \textbf{all} & \textbf{all} \\
        \midrule
         & & & & 0.0631 & 0.1486 & 0.2313 & 0.2395 & 0.1142 & 0.1431 & 0.1932 & 0.1612 & 0.2350 \\
        \checkmark   &     &    &   & 0.0721 & 0.0995 & 0.1456 & 0.1904 & 0.0870 & 0.1277 & 0.1928 & 0.1304 & 0.2069   \\
        \checkmark   & \checkmark  &  &    & 0.0603 & 0.0848 & 0.1232 & 0.1715 & 0.0759 & 0.1084 & 0.1608 & 0.1118 & 0.1889 \\
        \checkmark   & \checkmark &  \checkmark & & 0.0558 & 0.0604 & 0.0903 & 0.1679 & 0.0620 & 0.1091 & 0.1834 & 0.1039 & 0.1791 \\
        \midrule
        \checkmark  & \checkmark &  \checkmark &  \checkmark & 0.0570 & 0.0669 & 0.0948 & 0.1467 & 0.0670 & 0.0820 & 0.1084 & 0.0887 & 0.1695 \\
        \bottomrule
    \end{tabular}
    \caption{Overall ablation for the effectiveness of each component.}
    \label{tab:abl_all}
\end{table*}

\section{Ablations}

We demonstrate the effectiveness of each proposed component in~\cref{tab:abl_all}.
We use the architecture of \textit{VoxelNext}~\cite{chen2023voxenext} as the baseline model, then gradually update the baseline model with the proposed components. We start with \textit{VoxelNext} for two reasons: 1) it is one of the state-of-the-art point cloud object detection models with a fully sparse architecture design, and 2) it provides a good balance between computational costs and performance. We report the MPJPE for our ablations.
The results indicate all the individual components can contribute to improving keypoint estimation. Next, we further present the ablation studies to show the alternative design choices of the individual component.

\noindent\textbf{Spatially Aware BEV}
The use of the BEV representation significantly simplifies the detection problem by collapsing the 3D voxel space into a 2D feature map. This ablation evaluates the effectiveness of the proposed spatially aware BEV module. We first evaluate the direct use of a na\"ive 3D representations, followed by experiments with the spatially aware BEV. The findings, as shown in~\cref{tab:abl_bev}, indicate that our spatially aware BEV yields superior performance. The direct deployment of the 3D representation results in severe overfitting and, therefore, low performance. In addition, we also show that increasing the number of channels during the BEV projection can effectively improve the model performances, by compensating for information loss during projection. Overall, our spatially aware BEV strikes a balance that retains spatial acuity beyond basic BEV for resolving keypoint relationships while avoiding the complexity of full 3D convolutions.

\begin{table}[h]
    \scriptsize
    \setlength\tabcolsep{2.5pt}
    \centering
    \begin{tabular}{l|c|ccccccc|c}
        \toprule
         & cp. & head & shoulders & elbows & wrists & hips & knees & ankles & all \\
        \midrule
        3D & - & 2.4620 & 2.4559 & 2.4492 & 2.4449 & 2.4394 & 2.4264 & 2.419 & 2.4422 \\
        Ours & \xmark & 0.0688 & 0.0714 & 0.0982 & 0.1657 & 0.0723 & 0.1029 & 0.1595 & 0.1053 \\
        Ours & \cmark & 0.0570 & 0.0669 & 0.0948 & 0.1467 & 0.0670 & 0.0820 & 0.1084 & 0.0887 \\
        \bottomrule
    \end{tabular}
    \caption{Ablation study for the spatially aware BEV module. \textit{Cp.} denotes if to expand the number of channels to compensate for the information loss during the 2D projection.}
    \label{tab:abl_bev}
\end{table}

\noindent\textbf{Different Attention Mechanism}
\label{sec:abl-attn}
This ablation study assesses the effectiveness of the box-attention mechanism within point cloud processing. Recent advancements, such as the stratified self-attention~\cite{lai2022stratified}, focus on aggregating long-range contextual information, particularly beneficial for segmentation tasks. However, for keypoint estimation tasks, capturing global dependencies is less crucial. Instead, our approach utilizes local box-attention, which concentrates on adjacent local regions. The results, as presented in~\cref{tab:abl_attn}, demonstrate that local box-attention outperforms other methods. Interestingly, we found that the stratified attention mechanism could slightly impair performance. We suspect that the box-based approach concentrates on areas most relevant to each keypoint location, whereas long-range attention may cause the network to overlook local, dense details. As a result, the box-based attention mechanism allows efficient modeling of local keypoint distributions, without excessive computation or over-smoothing from global aggregation.

\begin{table}[h]
    \scriptsize
    \setlength\tabcolsep{1.8pt}
    \centering
    \begin{tabular}{l|cccccccc}
        \toprule
         & head & shoulders & elbows & wrists & hips & knees & ankles & all \\ 
        \midrule
        w/o & 0.0659 & 0.0956 & 0.1405 & 0.1855 & 0.0831 & 0.1077 & 0.1515 & 0.1181 \\
        stratified  & 0.0650 & 0.0911 & 0.1347 & 0.1995 & 0.0819 & 0.1245 & 0.1919 & 0.1266\\
        box & 0.0570 & 0.0669 & 0.0948 & 0.1467 & 0.0670 & 0.0820 & 0.1084 & 0.0887 \\
        \bottomrule
    \end{tabular}
    \caption{Different self-attention methods. \textit{w/o} denotes no attention applied.}
    \label{tab:abl_attn}
\end{table}



\section{Conclusion}

In this work, we proposed a new 3D fully sparse neural network for estimating dense human poses from point clouds. Our method combines several novel components including sparse selective kernel layers, box-attention layers, spatially aware multi-scale BEV fusion, and hybrid feature learning to accurately predict human body keypoints. Experiments on the Waymo dataset demonstrate the advantages of our approach compared to prior art and we demonstrate improved performance compared to other approaches trained on the same data as well as other approaches trained with additional data.

Despite these advancements, we further identify certain areas for future exploration and improvement. As mentioned above, this work used a small volume of training data, but it could benefit from a larger-scale dataset. While we focus on single-frame point clouds, future work could leverage temporal information across sequences of LiDAR point clouds. Additionally, instead of the straightforward estimation of keypoints, future work may adopt inverse kinematics to include physical constraints on human body movement.
Aside from refining estimated keypoint locations, this may especially be useful to handle real-world challenges such as occlusion within motion.

{
    \small
    \bibliographystyle{ieeenat_fullname}
    \bibliography{main}
}

\input{sec/X_suppl}

\end{document}

%% file: preamble.tex
\usepackage[dvipsnames,table]{xcolor}


%% file: sec/X_suppl.tex
\clearpage
\appendix
\setcounter{page}{1}
\maketitlesupplementary

\section{Technical Details}
\label{sec:tech_details}

This section presents additional technical details of the network, loss functions, and metrics used.

\subsection{Supplementary Network Details}
\label{sec:suppl_net}

\begin{figure}[h]
    \centering
    \includegraphics[width=.9\linewidth]{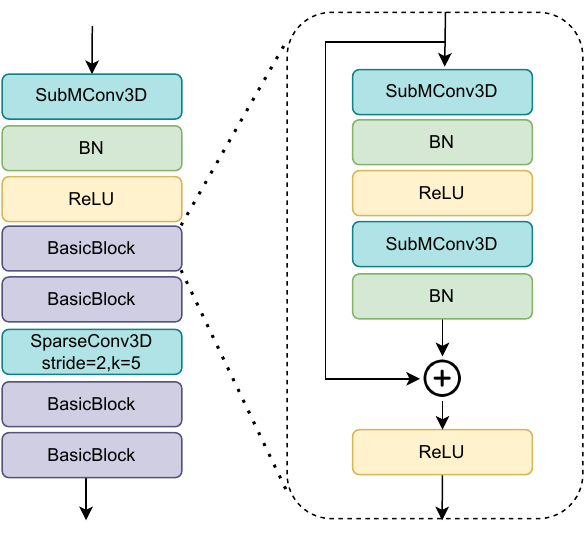}
    \caption{The architecture of the stem module.}
    \label{fig:stem}
\end{figure}

\begin{figure}[h]
    \centering
    \includegraphics[width=.28\linewidth]{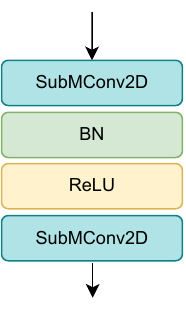}
    \caption{The architecture of prediction heads.}
    \label{fig:head}
\end{figure}

The architecture of the stem module and prediction heads are presented in~\cref{fig:stem,fig:head}. The stem module includes CONV-BN-ReLU blocks with skip connections to extract low-level features. It contains one downsampling layer to obtain a smaller feature map.
The model uses seven prediction heads. These heads predict: 1) the size of the bounding box, 2) the rotation of the bounding box, 3-5) the location of the box center and keypoints along the x, y, and z axes, 6) the visibility of keypoints, and 7) the Intersection over Union (IoU). Notably, we incorporate the IoU prediction to enhance performance, following~\cite{hu2022afdetv2}.


\subsection{Losses}
\label{sec:loss}

We use three types of losses in our work including the skeleton loss. Notably, the ground truth annotations are converted into the same sparse representation as the predictions for loss computation.

\textbf{Heatmap Loss}
Our network outputs a set of heatmaps, one per class. This heatmap encoding allows our model to classify and localize objects in 3D space simultaneously.
In the training phase, we assign positive heatmap indices based on ground truth annotations. Specifically, we identify the voxel closest to the annotated bounding box center and mark that voxel with a positive heatmap value.
We supervise these heatmaps using an adapted focal loss function~\cite{lin2017focal,law2018cornernet,chen2023voxenext}. With the annotated and predicted heatmaps $I$ and $\hat{I}$, we have:
\begin{equation}
\footnotesize
    FL(I, \hat{I}) = \frac{-1}{N} \sum_{c=1}^C \sum_{v=1}^V 
\begin{cases}
    (1-\hat{I})^\alpha \cdot \log(\hat{I}), & \text{if } I = 1\\
    \log{(1-\hat{I})} \cdot \hat{I}^\alpha \cdot (1 - I)^\beta,              & \text{otherwise}
\end{cases},
\end{equation}
where $N$, $C$, $V$ are the batch size, number of channels, and number of voxels, respectively. $\alpha$ and $\beta$ are the hyper-parameters to weigh each voxel. We use $\alpha=2$ and $\beta=4$ in this work, following~\cite{law2018cornernet}.

\textbf{L1 Regression Loss}
We adopt a simple L1 loss for other prediction heads of coordinates and keypoint visibilities. With the ground truth and predicted values $Y$ and $\hat{Y}$, we have:
\begin{equation}
    L1(Y, \hat{Y}) = \frac{1}{N} \sum_{c=1}^C ||Y - \hat{Y}||_1.
\end{equation}

\begin{figure*}[!b]
    \centering
    \includegraphics[width=0.95\textwidth]{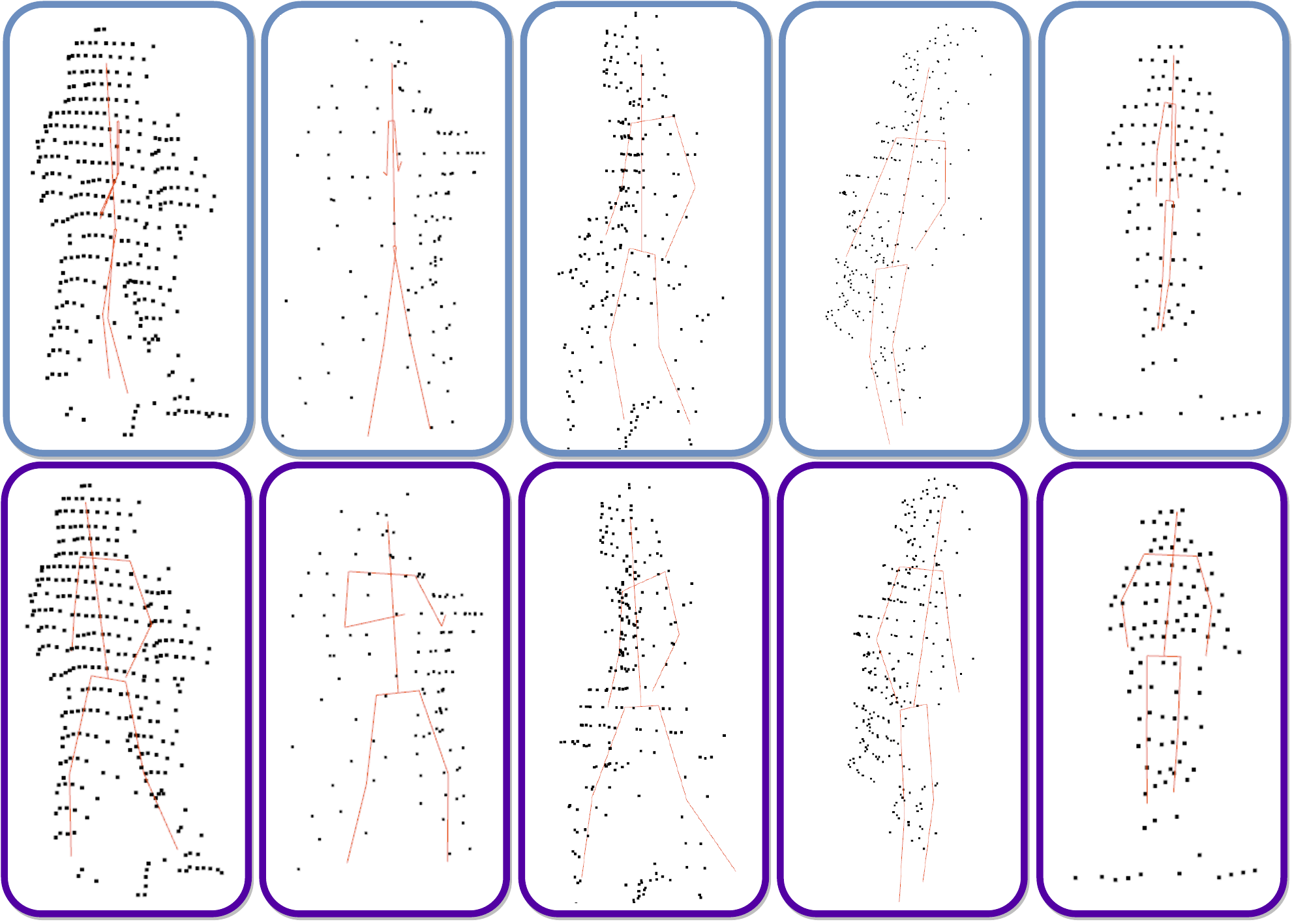}
    \caption{A visual demonstration of the baseline model (top row) and the proposed \textit{VoxelKP} (bottom row) on matched human objects.}
    \label{fig:visual_sup1}
\end{figure*}

\textbf{Skeleton Regularization}
We propose to use a skeleton loss to encode prior information about the relative positioning of keypoints. For this purpose, we include bone length regularization in the loss function. This term computes the distance between the ground truth bone length and the predicted bone length.
Specifically, given the ground truth keypoint locations $Y$ and predicted keypoint locations $\hat{Y}$, we first compute the skeleton bone lengths $BL(Y)$ and $BL(\hat{Y})$ by calculating the Euclidean distance between connected keypoint pairs.
The skeleton loss is then calculated as the Huber loss $h(\cdot)$ between the predicted bone lengths $BL(\hat{Y})$ and ground truth bone lengths $BL({Y})$, resulting in:
\begin{equation}
    \text{SK}(\hat{Y},Y)=h(BL(\hat{Y}), BL({Y})).
\end{equation}
This enforces the model to predict keypoint locations that respect the biomechanical constraints of bone lengths in the human skeleton. Matching the distribution of predicted bone lengths to the ground truth, ensures awareness of the spatial relationships between different joints. The skeleton loss penalizes predicted keypoints that violate the physical constraints of bone lengths, acting as a strong prior for plausible human poses.

\begin{figure*}
    \centering
    \includegraphics[width=0.95\textwidth]{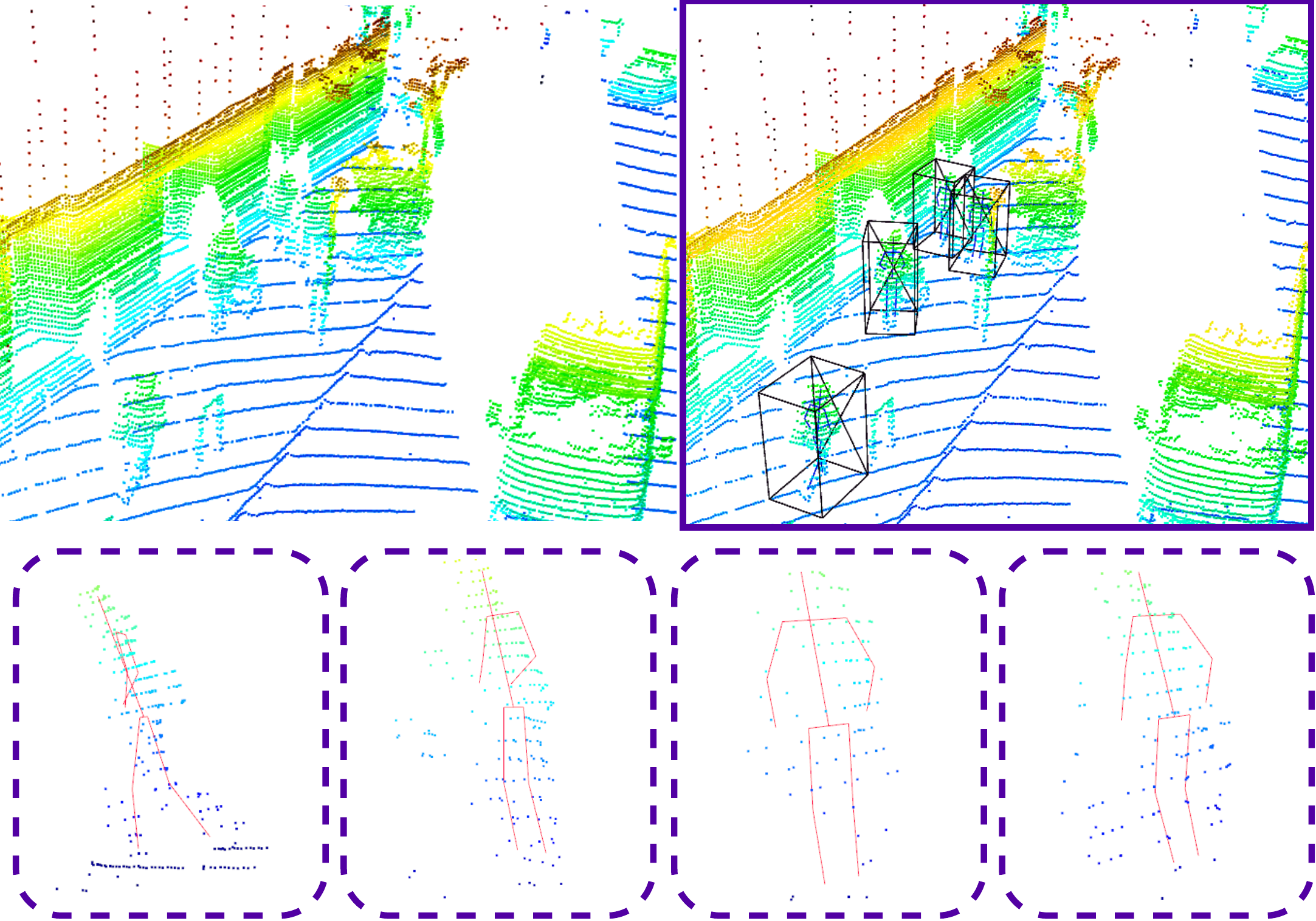}
    \caption{A visual demonstration of the baseline model (top left) and the proposed \textit{VoxelKP} (top right). Our method detects the human objects that the baseline method fails to. The bottom line shows object-level keypoint localization performance.}
    \label{fig:visual_sup2}
\end{figure*}

\subsection{Metrics}
\label{sec:metrics}

We use mean per-joint position error (MPJPE), pose estimation metric (PEM), and object keypoint similarity (OKS) to evaluate our method. Formally, let $\hat{Y} \in \mathbb{R}^{J\times 3}$ be the predicted keypoints of a human, $Y \in \mathbb{R}^{J\times 3}$ be the ground truth, and $v_j \in {0,1}$ be the visibility of each joint $j$. The MPJPE metric is defined as:
\begin{equation}
    \text{MPJPE}(Y,\hat{Y}) = \frac{1}{\sum_j v_j}\sum_{j\in [J]} v_j||y_j -\hat{y}_j||_2.
\end{equation}
Note that MPJPE requires a one-to-one match between the keypoints predictions and ground truth. Therefore, a Hungarian matching is performed to match the predicted and annotated keypoints before calculating the MPJPE.

PEM further takes into account the matching accuracy that is essentially a sum of the MPJPE over visible matched keypoints with a penalty for unmatched keypoints. Note that the unmatched keypoints include both the ground truth keypoints without matching predicted keypoints and the predicted keypoints without matching ground truth objects.
\begin{equation}
    \text{PEM}(Y,\hat{Y})=\frac{\sum_{i\in M}||y_j-\hat{y}_j||_2+C|U|}{|M|+|U|},
\end{equation}
where $M$ is a set of indices of matched keypoints, $|U|$ is a set of indices of unmatched keypoints, and $C=0.25$ is a constant penalty for an unmatched keypoint.

Additionally, we include the classic metric of OKS in this work.
The OKS metric is not computed per keypoint, it is a relative metric computed for each human body.
In OKS, each ground truth object also has a scale s which we define as the square root of the object segment area.
OKS is computed as the arithmetic average across all labeled keypoints in an instance.
\begin{equation}
    \text{OKS}=\frac{\sum_j e^{-\frac{d_j^2}{2s^2k_j^2}v_j}}{\sum_i v_j}
\end{equation}
where $d_j$ is the Euclidean distance between each corresponding ground truth and detected keypoint, $k_j$ is a per-joint constant provided by COCO~\cite{lin2014microsoft}. The reported OKS@KP is averaged over multiple OKS values, which are calculated for OKS thresholds starting at $0.50$, increasing in steps of $0.05$, and ending at $0.95$.

\section{Visual Results}

We present additional visual results in this section. Majorly, we show our method locates the keypoints with better precision in~\cref{fig:visual_sup1}, and can detect the human objects better than the baseline in~\cref{fig:visual_sup2}.


